%% file: pprai.tex
\documentclass{pprai}

\usepackage[utf8]{inputenc}
\usepackage[T1]{fontenc}
\usepackage{graphicx}
\usepackage{amsthm}
\usepackage{amsmath}
\usepackage{bbm}
\usepackage{txfonts}
\usepackage{url}
\usepackage{algpseudocode}
\usepackage{placeins}

\usepackage[linesnumbered,ruled]{algorithm2e}

\SetKwRepeat{Do}{do}{while}
\DeclareMathOperator{\argmin}{arg\,min} 

\usepackage{background}
\backgroundsetup{
  position=current page.north,
  angle=0,
  nodeanchor=north,
  vshift=-5mm,
  opacity=1,
  scale=1,
  contents=Pre-print: The full text will be published in the proceedings of PP-RAI 2024.
}

\title{A-PETE: Adaptive Prototype Explanations of Tree Ensembles}
\headtitle{Prototype Explanations of a Tree Ensemble}


\author{Jacek Karolczak$^{[0000-0001-5414-960X]}$, Jerzy Stefanowski$^{[0000-0002-4949-8271]}$}
\headauthor{J. Karolczak, J. Stefanowski}
\affiliation{%
  Poznan University of Technology\\
  ul. Piotrowo 3, 60-965 Poznan, Poland\\
jacek.karolczak@student.put.poznan.pl  jstefanowski@cs.put.poznan.pl}


\begin{document}

\maketitle

\begin{abstract}
The need for interpreting machine learning models is addressed through prototype explanations within the context of tree ensembles. An algorithm named Adaptive Prototype Explanations of Tree Ensembles (A-PETE) is proposed to automatise the selection of prototypes for these classifiers. Its unique characteristics is using  a specialised distance measure and a modified k-medoid approach. Experiments demonstrated its competitive predictive accuracy with respect to earlier explanation algorithms. It also provides a a sufficient number of prototypes for the purpose of interpreting the random forest classifier.
\end{abstract}

\section{Introduction}

In recent years, there has been a growing need to explain  machine learning black box models, which has led to the development of the field of explainable Artificial Intelligence (XAI) \cite{bodria-pisa}. Making machine learning models' decisions interpretable, enables users to trust and rely on models' predictions. In general, one can distinguish between \textit{global} and \textit{local explanations}. The global explanations refer to studying more comprehensive interactions between input attributes and output, target variables for the complete data set, rather aimed at a general explanation of how ML model works. On the other hand, the local explanation attempts to discover the reasons for a single decision of the model for a given instance. \cite{bodria-pisa}.

While explanations can be given in different forms, this study focuses on explanations articulated via \textit{prototypes}. As defined in~\cite{Molnar} prototypes are certain instances (representative real examples or synthetic ones) defined on the basis of the learning dataset to explain the behaviour of machine learning models, or sometimes also to explain the underlying data distribution. A user can comprehend the model’s reasoning by comparing prototypes to user's data. Having a well identified set of prototypes, an application-specific explanation can be yielded more easily than more complex explanations. For instance, prototypes can act as a global interpretation of the underlying behaviour of an obscure black box ML model, by presenting to the user the limited number of representative prototypes for a class. Prototypes can also be used for local explanation of a single decision, by showing the similarity of the predicted instance to the nearest or max few nearest representative exemplar(s) -- prototype(s) \cite{obermair_proto_or_example}. Local explanation can be further enhanced by showing the most similar prototypes belonging to other classes, to give deeper insight into the decision.

Inspired by Tan et al~\cite{tan_xai_prototype_tree_space}, we will consider the new idea  of finding such prototypes for explaining Random Forest predictions, which uses a specialised distance measure and a modified k-medoids algorithm. However, it is important to identify a limited number of such prototypes due to the perceptual abilities of the recipient, which was not always preserved in experiments~\cite{tan_xai_prototype_tree_space}. Furthermore, we believe that the user should be supported in this task. Therefore, the main goal of our paper includes studying the automatic selection of the number of these prototypes. The usefulness of this  algorithm (named A-PETE) and earlier proposals are experimentally evaluated.

\section{The distance function for tree ensembles}

\noindent Our work extends the proposal of a similarity measure for instances within a tree ensemble \cite{tan_xai_prototype_tree_space}, which considers predictions and distributions of learning examples in the tree leaves. The proposed approach works under the condition of equal contribution of each tree to the final decision.
Let $t$ represent the number of trees in the tree ensemble (TE). The $i$-th tree ($i \in [t]$) partitions the feature space into regions $R_{i,j }$, each corresponding to a leaf $\tau_{i,j}$. Each tree induces an individual classifier assigning each point $x \in X$ to a single region $R_{i, j}$:

\begin{equation}
    c_i^{\text{Tree}}(x) = \sum_{j=1}^{\tau_i} \alpha_{i, j} \mathbbm{1}(x \in R_{i, j}) \,,
    \label{eq:tree}
\end{equation}

\noindent where $\alpha_{i,j}$ is the predicted value in the $j$-th leaf of the $i$-th tree. $\mathbbm{1}$ denotes the indicator function. The tree ensemble classifier is the average over all trees:

\begin{equation}
    c^{\text{TE}}(x) = \frac{1}{t}\sum_{i = 1}^t c_{i}^{\text{Tree}} = \frac{1}{t}\sum_{i = 1}^t \sum_{j=1}^{\tau_i} \alpha_{i, j} \mathbbm{1} (x \in R_{i, j}) \,.
    \label{eq:ensemble}
\end{equation}

Thus, the proximity of two instances $x_1$ and $x_2$ is given as the mean number of trees in which both instances land in the same leaf and can be expressed as

\begin{equation}
    \text{proximity}^{\text{TE}}(x_1, x_2) = \frac{1}{t}\sum_{i = 1}^t \sum_{j=1}^{\tau_i} \mathbbm{1}(x_1 \in R_{i, j})\mathbbm{1}(x_2 \in R_{i, j})\,.
    \label{eq:proximity}
\end{equation}

The distance metric can be derived from the proximity function $d^{\text{TE}}(x_1, x_2) =  1 - \text{proximity}^{\text{TE}}(x_1, x_2)$.

Given that the regions $\{R_{i, j}\}_{j=1}^{\tau_i}$ form a partition of the feature space, each point $x \in X$ belongs to at most one region. The inner sum in the proximity definition yields values of either 0 or 1 for each tree. Thus, the proximity, a convex combination of all trees, and consequently the distance, also lies within the range~of~[0,~1].

\section{Prototype selection as medoids}

\noindent The set of discovered prototypes $P$ should satisfy the following  properties:
\begin{itemize}
    \item Reality --  $P$ consists of real objects from the learning set ($P \subset X$).
    \item Coverage -- For a class $c$, the prototypes in $P^c$ should collectively span the class, ensuring that each instance $x^c \in X^c$ is represented by (close to) a prototype in $P^c$ ($P^c = \argmin_{p^c \in X^c} \sum_{i=1}^{|P^c|} \sum_{j=1}^{|X^c|} \text{distance}^{\text{TE}}(p_i^c, x_j^c)$).
    \item Compactness -- The set of prototypes should be the minimal collection of instances that satisfies the aforementioned conditions ($|P| \ll |X|$).
\end{itemize}

This work is inspired by the work of Tan et al~\cite{tan_xai_prototype_tree_space}, where the K-medoids problem was adapted to the task of prototype selection, that is the objective is to find a subset of examples $P \subset X$, where $|P|=k$. However, a drawback of \cite{tan_xai_prototype_tree_space} is that the user has to define the number of medoids (prototypes) $k$.

We propose the Adaptive Prototype Explanations of Tree Ensembles (A-PETE), which automatically selects $k$ prototypes.  We adopt the greedy submodular prototype selection algorithm (SM-A), which  minimises the function:
\begin{equation}
    f(P) = \sum_{c}^{C} \sum_{i}^{|X^c|} \min_{p^c \in {P^c}} d(x_i^c, p^c)\,.
    \label{eq:func-f}
\end{equation}

The A-PETE algorithm is composed of three parts. Initially, a set of phantom prototypes $P'$ is created, where the distance between these prototypes and all $x \in X$ equals 1. Additionally, an empty set is prepared for real prototypes~$P$. The core of the algorithm lies in the iterative prototype selection process. In each iteration, the algorithm identifies an instance $x^{*}$ that maximises the difference between the total distance of phantom prototypes $P'$ and the set is a union of phantom prototypes $P'$, prototypes $P$ and the maximising instance $x^{*}$. The maximising instance $x^{*}$ is added to the set of prototypes $P$. The algorithm computes the change in the objective function $\Delta$ resulting from adding the selected instance $x^{*}$ to the set of prototypes.

The main novelty is that our algorithm maintains the difference $\Delta$ between consecutive objective function changes to monitor the progress. If the relative change surpasses the threshold, the algorithm concludes, and the set of prototypes $P$ is the final output. A comprehensive pseudocode is presented as Algorithm \ref{alg:A-PETE}.

\begin{algorithm}[t]
    \SetKwInOut{Input}{Input}
    \SetKwInOut{Output}{Output}

    \Input{Set of points $X$, distance function $d : X^2 \mapsto [0, 1]$, class assignment $c : X \mapsto [q]$, control parameter $\alpha \in (0, 1)$}
    \Output{Set of prototypes $P$}
    Create set of phantom exemplars $P' = \{p_1', ..., p_q'\}$ and set $d(p_i', x) = d(x, p_i') = 1$ for all $x \in X$ \\
    $\Delta \gets 0$ \\
    P $\gets \varnothing$ \\
    \While{True}{
        $x^{*} \gets \underset{x \in X}{\arg\max}[f(P') - f(P' \cup P \cup \{x\})]$ \\
        $\Delta' \gets f(P' \cup P \cup \{x^{*}\}) - f(P' \cup P) $ \\
        $P \gets P \cup \{x^{*}\}$ \\
        \If{
            $\frac{|\Delta - \Delta'|}{\Delta'} < \alpha$
        }{
            break
        }{
            $\Delta' \gets \Delta$ \\
        }
    }
    \caption{Adaptive Prototype Explanations of Tree Ensembles  (A-PETE).}
    \label{alg:A-PETE}
\end{algorithm}

\section{Experimental evaluation}

We will compare our proposal A-PETE against the earlier two proposals of selecting prototypes (SM-A and SM-WA) coming from \cite{tan_xai_prototype_tree_space}. 

The SM-A, SM-WA, and A-PETE were implemented on top of the Random Forest ensemble with scikit-learn Python library\footnote{https://scikit-learn.org}. To be consistent with hyper-parameters from~\cite{tan_xai_prototype_tree_space}, the random forests were trained with 1000 trees without limiting the tree depth. The number of features was tuned from values: $\sqrt{p}$, $7$, $0.33p$, $0.5p$, $0.7p$, where $p$ is the number of features. The values of $p$ yielding the best results, found during hyperparameter tuning, for each algorithm is presented in Table~\ref{tab:comparison-p}.

\begingroup
\setlength{\tabcolsep}{3pt}
\begin{table}[t]
    \centering
    \caption{The values of $p$ (the number of features in nodes of Random Forest)  yielding best results (Table \ref{tab:comparison-results}), found in hyperparameter tuning.}
    \begin{tabular}{*{7}{c}}
        \hline
        & Breastcancer & Diabetes & Compass & RHC & Mnist & Caltech256 \\
        \hline \hline
        SM-A & sqrt & 0.33 & 0.33 & 7 & sqrt & sqrt \\
        SM-WA & sqrt & sqrt & 0.5 & 0.5 & sqrt & 0.33 \\
        A-PETE & sqrt & 0.33 & 0.5 & 7 & 0.33 & sqrt \\
        \hline
    \end{tabular}
    \label{tab:comparison-p}
\end{table}
\endgroup

The SM-A and SM-WA algorithms were executed with a search scope of up to 20 prototypes. This choice was guided by our empirical study revealing that a higher number of prototypes may pose challenges for interpretation due to inherent limitations in cognitive capabilities. However, it should be noted that SM-A and SM-WA algorithms could produce a much higher number of final prototypes (see the original results published in \cite{tan_xai_prototype_tree_space}).

We chose 6 popular datasets, often used in the related studies (in particular also in \cite{tan_xai_prototype_tree_space}). These data  have different sizes and characteristics. Due to pages limits we only gave their names in Table \ref{tab:comparison-results} and the reader can find  more details about them. Note that for CALTECH-256, two easily confused classes guitar and mandolin were selected and deep features were extracted by ResNet-50 trained on ImageNet. For Mnist digits 4 and 9 as look-alike classes were selected and raw pixel values were used as features. Compass contains several discriminative features, so it was preprocessed following the strategy proposed in AIF-360\footnote{https://github.com/Trusted-AI/AIF360/}. All categorical features of RHC and Compass underwent an ordinal encoding. 

Having no real user case study, we decided to use a surrogate model approach to assessing the predictive ability, i.e. the fidelity measure in relation to the Random Forest prediction. For all the prototype selection algorithms, the test example is assigned to the class of the nearest prototype. Such an assessment procedure is consistent with Tan et al.~\cite{tan_xai_prototype_tree_space}. For Mnist the original train-valid-test split was maintained. For the remaining datasets, the train-valid-test split was done with proportions 60\%-20\%-20\% and class stratification.

The A-PETE approach was compared to the adaptive greedy submodular (SM-A) and weighted adaptive greedy submodular (SM-WA) introduced in \cite{tan_xai_prototype_tree_space} using weighted -- balanced accuracy as the evaluation metric suitable to deal with class imbalances~\cite{Brzezinski2018}. It is computed as a weighted sum of the accuracy of each class, where the weight of each class is determined by dividing the number of instances of that class by the total number of instances in the dataset. It is noteworthy that A-PETE was executed for Breastcancer, Diabetes, Mnist, and Caltech256, utilising the default $\alpha = 0.05$. Due to sub-optimal outcomes for Compass and RHC, A-PETE was subject to additional evaluation with $\alpha = 0.01$.

\section{Results}

\begingroup
\setlength{\tabcolsep}{3pt}
\begin{table}[t]
    \centering
    \caption{The best weighted accuracy achieved using Random Forest (RF) and 1-NN run on prototypes selected using adaptive greedy submodular prototype selection (SM-A), weighted adaptive greedy submodular prototype selection (SM-WA), and Adaptive Prototype Explanations of Tree Ensembles (A-PETE). The number of prototypes in parentheses.}
    \begin{tabular}{*{7}{c}}
        \hline
        & Breastcancer & Diabetes & Compass & RHC & Mnist & Caltech256 \\
        \hline \hline
        RF & 0.93 & 0.73 & 0.66 & 0.75 & 0.99 & 0.69 \\
        SM-A & 0.92 (8) & 0.74 (3) & 0.30 (20) & 0.74 (12) & 0.97 (14) & 0.70 (16) \\
        SM-WA & 0.92 (8) & 0.72 (2) & 0.30 (20) & 0.40 (10) &  0.97 (11) & 0.72 (5) \\
        A-PETE & 0.92 (7) & 0.73 (5) & 0.32 (23) & 0.73 (9) & 0.97 (19) & 0.70 (6) \\
        \hline
    \end{tabular}
    \label{tab:comparison-results}
\end{table}
\endgroup

The experiment highlighted the effectiveness of automatically selecting prototypes. In terms of weighted (balanced) accuracy, the surrogate (nearest prototype) model achieved  almost the same  predictive performance for both SM-A and A-PETE on half of the datasets. For the Diabetes and RHC datasets, A-PETE provided a weighted accuracy only one percentage point lower than SM-A. In contrast, A-PETE outperformed SM-A by two percentage points in the case of the Compass dataset, indicating the negligibility of the difference. Except for Compass, all three algorithms achieved accuracy levels comparable to the original Random Forest ensemble, which suggests that prototypes may capture essential information for imitating the Random Forest's decision-making process.

Moreover, A-PETE consistently approached the number of prototypes from SM-A without sacrificing predictive performance, suggesting sufficient prototype selection. In contrast, SM-A still requires manual tuning of the optimal~$k$ value.

To sum up, A-PETE offers a promising solution for the automatic prototype selection in tree ensembles, addressing the need for the local and global explanation of the tree ensembles.

\noindent{\textbf{Ack}: The research on this paper has been partially supported by PUT SBAD grant.}

\bibliography{pprai}
\bibliographystyle{pprai}

\end{document}